\documentclass[9pt,shortpaper,twoside,web]{ieeecolor2}
\usepackage{generic}
\usepackage{cite}
\usepackage{amsmath,amssymb,amsfonts}
\usepackage{algorithmic}
\usepackage{graphicx}
\usepackage{textcomp}
\usepackage{float}
\usepackage{booktabs}
\usepackage{caption}
\usepackage{multirow}
\usepackage[section]{placeins}
\usepackage{threeparttable} 
\usepackage[colorlinks=true,
linkcolor=blue,
citecolor=blue,
urlcolor=blue]{hyperref}  

\begin{document}
\title{A Continuous sEMG-Based Prosthetic Hand Control System Without Motion or Force Sensors}
\author{
	Gang Liu\textsuperscript{1,2},
	Ye Sun\textsuperscript{1,2},
	Zhenxiang Wang\textsuperscript{2,3},
	Chuanmei Xi\textsuperscript{1,2},
	Ziyang He\textsuperscript{4,*},
	Shanshan Guo\textsuperscript{5},
	Rui Zhang\textsuperscript{1,2,*},
	and Dezhong Yao\textsuperscript{6,7,*}
	\thanks{Gang Liu. Email: gangliu\_@zzu.edu.cn, Dezhong Yao. Email: dyao@uestc.edu.cn, Rui Zhang. Email: ruizhang@zzu.edu.cn, Ziyang He. Email: zyhe@zzu.edu.cn}  
	\thanks{This work is supported by the National Natural Science Foundation of China (62303423), the STI 2030-Major Project (2022ZD0208500), Postdoctoral Science Foundation of China (2024T170844, 2023M733245), the Henan Province key research and development and promotion of special projects (242102311239), Shaanxi Province key research and development plan (2023GXLH-012), the Shanghai Key Laboratory of brain-computer Collaborative Information Behavior (2023KFKT005).}
	\thanks{\textsuperscript{1}School of Electrical and Information Engineering, Zhengzhou University, Zhengzhou 450001, China}
	\thanks{\textsuperscript{2}Henan Key Laboratory of Brain Science and Brain Computer Interface Technology, Zhengzhou University, Zhengzhou 450001, China}
	\thanks{\textsuperscript{3}Engineering Department of International College, Zhengzhou University, Zhengzhou 450001, China}
	\thanks{\textsuperscript{4}School of Cyber Science and Engineering, Zhengzhou University, Zhengzhou 450000, China}
	\thanks{\textsuperscript{5}Shanghai Key Laboratory of Brain-Machine Intelligence for Information Behavior, Shanghai International Studies University, Shanghai, China}
	\thanks{\textsuperscript{6}Clinical Hospital of Chengdu Brain Science Institute, MOE Key Laboratory for NeuroInformation, University of Electronic Science and Technology of China, Chengdu, China}
	\thanks{\textsuperscript{7}Research Unit of NeuroInformation, Chinese Academy of Medical Sciences, Chengdu, China}
}
\maketitle

\begin{abstract}
Regressively-based surface electromyography (sEMG) prosthetics are widely used for their ability to continuously convert muscle activity into finger force and motion. However, they typically require additional kinematic or dynamic sensors, which increases complexity and limits practical application. To address this, this paper proposes a method based on the simplified near-linear relationship between sEMG and finger force, using the near-linear model ResDD proposed in this work. By applying the principle that a line can be determined by two points, we eliminate the need for complex sensor calibration. Specifically, by recording the sEMG during maximum finger flexion and extension, and assigning corresponding forces of 1 and -1, the ResDD model can fit the simplified relationship between sEMG signals and force, enabling continuous prediction and control of finger force and gestures. Offline experiments were conducted to evaluate the model’s classification accuracy and its ability to learn sufficient information. It uses interpolation analysis to open up the internal structure of the trained model and checks whether the fitted curve of the model conforms to the nearly linear relationship between sEMG and force. Finally, online control and sine wave tracking experiments were carried out to further verify the practicality of the proposed method. The results show that the method effectively extracts meaningful information from sEMG and accurately decodes them. The near-linear model sufficiently reflects the expected relationship between sEMG and finger force. Fitting this simplified near-linear relationship is adequate to achieve continuous and smooth control of finger force and gestures, confirming the feasibility and effectiveness of the proposed approach.
\end{abstract}

\begin{IEEEkeywords}
	Kinetic and kinematic sensor-free, Prosthetic hands, sEMG, Real-time systems 
\end{IEEEkeywords}

\section{Introduction}
\label{sec:introduction}

Hand amputees experience profound limitations in their daily lives, significantly impacting their independence and quality of life. Moreover, the loss of hand function also affects psychological well-being, leading to feelings of frustration, social isolation, and reduced self-esteem ~\cite{2021Quality}. The development of advanced prosthetic hands aims to address these limitations by providing amputees with greater dexterity control and sensory feedback. However, the lack of intuitive control interfaces and difficulty in performing simultaneous and coordinated movements are major obstacles hindering the widespread adoption of prosthetic hands currently ~\cite{Igual2019}. sEMG prosthetic hands enable intuitive and natural control by translating user intent into mechanical action using surface electromyography (sEMG) signals. The technological development of these devices relies on key advancements in areas such as signal processing, intent recognition algorithms, feedback systems, and human-machine interface design. This has positioned sEMG prosthetic hands as a highly promising approach to achieving intuitive prosthetic control ~\cite{Wu2010,10.4108/airo.7377}. Current applications of sEMG prosthetic hands primarily fall into two categories: Classification-based estimation of discrete gestures and regression-based estimation of continuous kinematics and kinetics.
\begin{figure*}[htbp]  
	\centering  
	\includegraphics[width=0.8\textwidth]{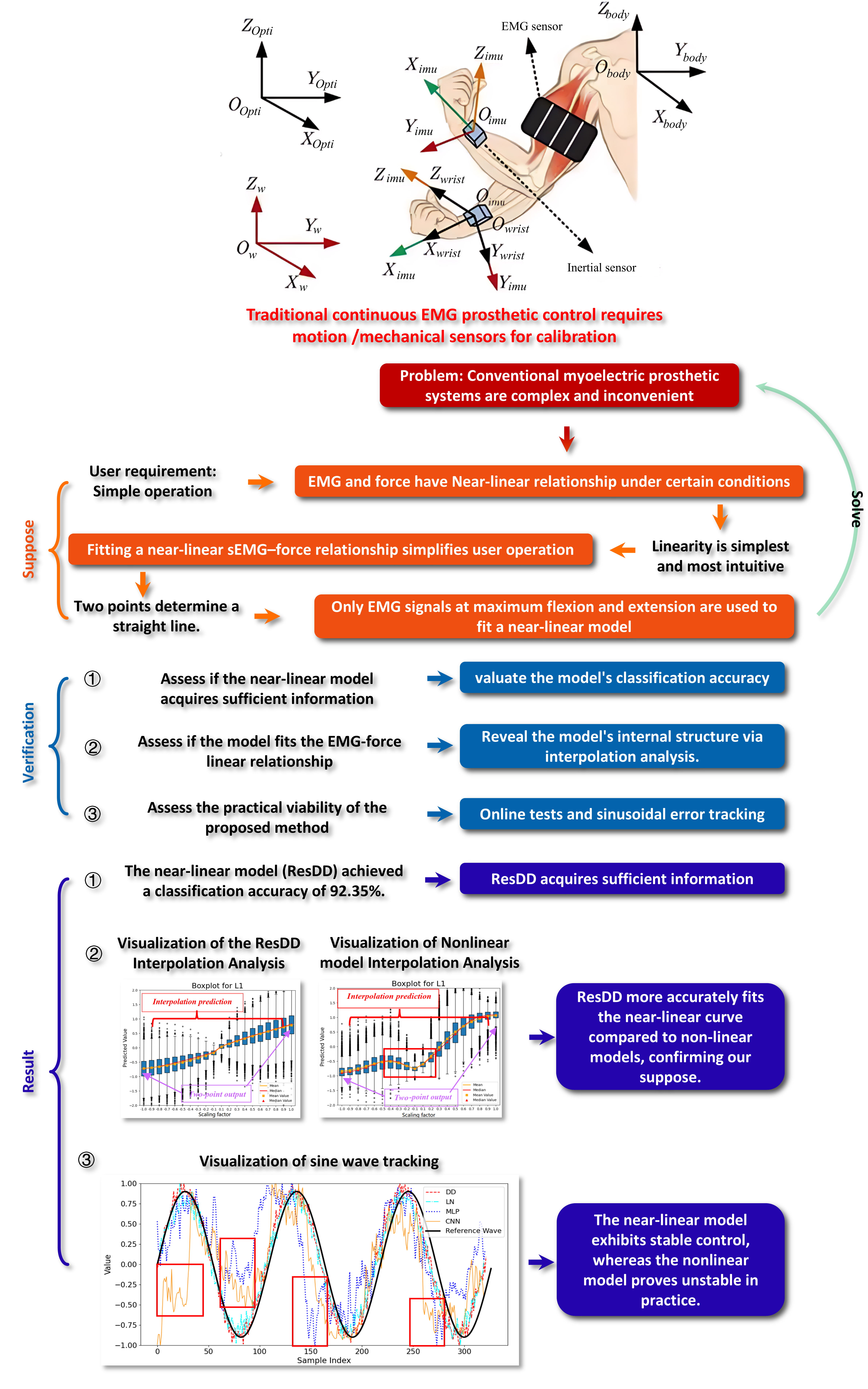} 
	\caption{Idea and verification process} 
	\label{fig:fig0}
\end{figure*} 
In the direction of gesture-based classification for sEMG prosthetic hand control, for example, Ulysse Côté-Allard et al. have aggregated sEMG data from multiple individuals to propose a classification method based on transfer learning ~\cite{CoteAllard2018}, while Hu, Y et al. have developed an attention-based hybrid CNN-RNN architecture that better captures the temporal characteristics of sEMG signals ~\cite{Hu2018}. Khosro Rezaee proposed a U-Net architecture incorporating a MobileNetV2 encoder and a novel Bidirectional Long Short-Term Memory (BiLSTM) optimized by a metaheuristic optimization to enhance spatial feature extraction in gesture and action recognition, Therese improving the accuracy of hand classification ~\cite{Rezaee2024}. However, the scope of application of classification methods is relatively limited, these methods only pay attention to the result of the movement, namely the formed gestures, but ignore the basic composition of the movement such as muscle connectivity to individual fingers force and velocity, which demonstrates high error rate in real-time control.

\begin{figure*}[htbp]  
	\centering  
	\includegraphics[width=0.8\textwidth]{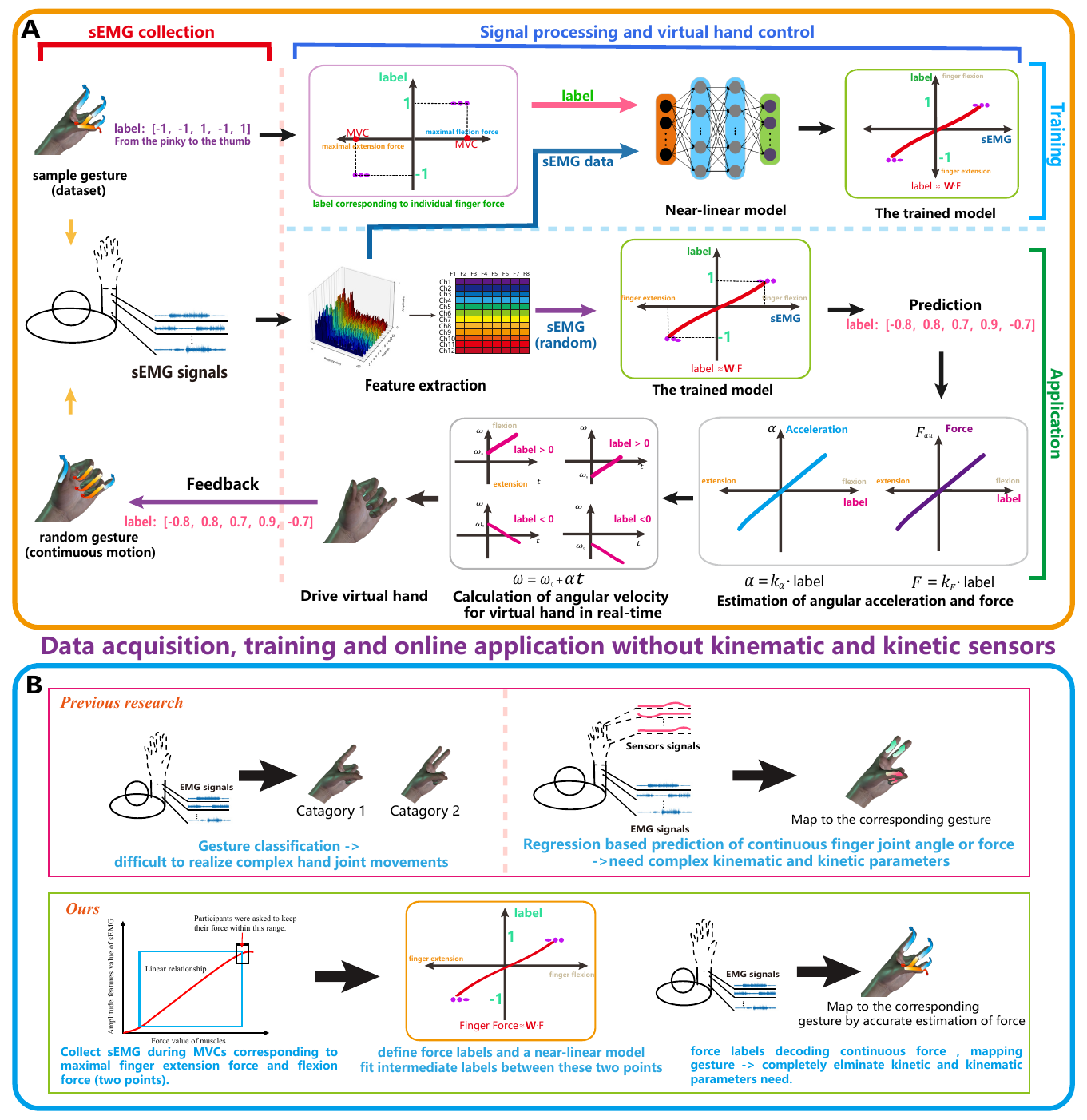} 
	\caption{Continuous prediction of finger force labels and virtual hands control system based on sEMG. Part A demonstrates the process of data acquisition, model training, and online application without the need for kinetic and kinematic sensors. Part B contrasts the differences in methods for gestures recognition and kinetic and kinematic information prediction between this study and previous research} 
	\label{fig:fig1}
\end{figure*} 

To address the limitations of classification-based methods, researchers have explored regression approaches to decode the fundamental processes of movement from sEMG. For instance, studies on finger force estimation, such as those by Hang Su et al. and our previous research have developed algorithms and models for human-robot interaction and continuous hand action decoding, respectively ~\cite{Su2021, Liu2021}. Similarly, Yang Zheng and Xiaogang Hu proposed a neural-driven approach for real-time finger force estimation based on MU discharge events from sEMG ~\cite{Zheng2019}. In the area of joint angle prediction, Zhang, Feng et al. employed a simple BP neural network to establish an $ m-$th order nonlinear model linking sEMG signals and human leg joint angles ~\cite{Zhang2012}, and Ding, Qichuan et al. proposed a state-space motion model with an unscented Kalman filter (UKF) for estimating multi-joint angles from sEMG  ~\cite{Ding2017}. Notably, Zhuo Wang introduced a Dual Transformer Network (DTN), aimed at simultaneously estimating multiple joint angles and torques from multi-channel sEMG signals of the lower limbs ~\cite{Wang2025}. Despite these advancements, the reliance on kinetic and kinematic sensors, as highlighted in ~\cite{Lv2017,Chai2021,Stapornchaisit2019,Chen2022}, increases system complexity and imposes stringent requirements for synchronized data collection, hindering the development of truly practical and user-friendly sEMG-based prosthetic systems.While significant progress has been made in predicting joint angles and estimating force and torque, existing methods often rely heavily on kinetic and kinematic sensors for data collection, which is inconvenient to use.

Research has shown that under certain conditions, the relationship between surface electromyographic (sEMG) and force can be simplified to near-linearity. This near-linearity not only simplifies the control system but also makes user operation more intuitive. In real-time applications, users can seamlessly integrate into a closed-loop control system, receiving feedback and making adjustments. To further avoid the cumbersome process of calibration based on kinematic or mechanical sensors, this paper, based on the simplified near-linear relationship between sEMG and force, proposes a simple continuous control myoelectric prosthetic hand system(The idea and verification process of this system is shown in \autoref{fig:fig0}). Specifically, the method simplifies the modelling process using the mathematical principle of "two points determine a line," eliminating complex sensor calibration. By recording sEMG during maximum finger flexion and extension and assigning corresponding forces of 1 and -1, the near-linearity model ResDD(see \autoref{fig:fig1}) effectively fits the relationship between sEMG and force, enabling continuous prediction and control of finger force and gestures. This paper first evaluates the model's classification accuracy through offline experiments to determine if it can effectively learn sufficient information. Additionally, use the interpolation analysis method to access the internal structure of the trained model, checking if the fitted curve aligns with the simplified near-linearity relationship between sEMG and force. Finally, online control and sine wave tracking experiments further validate the method's feasibility, especially demonstrating that the simplified near-linearity control curve helps users better receive feedback and respond in a closed-loop control system, resulting in more precise control. Experimental results show that the proposed method effectively extracts key information from surface sEMG and decodes them accurately. The near-linearity model better fits the expected relationship between sEMG and finger force. By fitting this simplified near-linearity relationship, the system achieves continuous, smooth control of finger force and gestures, demonstrating the method's feasibility and effectiveness. 

The main contributions of this paper are as follows:
\begin{enumerate}  
	\item This paper first proposes a prosthetic hand control system based on continuous sEMG that does not require motion or force sensors. By simplifying the near-linear relationship between sEMG and force and applying the mathematical principle of "two points determine a line," we simplify the modelling process, effectively addressing the dependence on motion/dynamics sensors in traditional regression-based myoelectric prosthetic systems.  
	
	\item This paper proposes the near-linear model ResDD, which retains the linearity while learning more information than traditional linear models. Comparison with linear models (LN) and nonlinear models (e.g. typical CNN and MLP) in fitting the relationship between sEMG and force, as well as real-time finger force control, shows that ResDD outperforms the others in fitting near-linear relationships and real-time decoding of sEMG for prosthetic control.
	
	\item Through online experiments, we validate that a simple control curve helps users better receive feedback and respond in a closed-loop control system, leading to more precise control.  
	
\end{enumerate}  

The rest of the paper is organized as follows: Section \ref{Methodology} describes Principle of the Proposed Method and model establishment, including data collection and preprocessing, model structure. Section \ref{Experimental setup} also describes the the model structure of three other models used for experimental comparative analysis with ResDD and corresponding experimental Settings. Our experimental results will be presented in Section \ref{Results}. Furthermore, our work is discussed in Section \ref{Discussion}. Finally, Section \ref{Conclusion} concludes this work.

\section{The near-linear method proposed in this paper}
\label{Methodology}

\subsection{Principle of the Proposed Method}
Previous studies have shown that, within a certain range, the force generated by muscle contraction exhibits an approximately linear relationship with the amplitude of the corresponding sEMG ~\cite{Dideriksen2010,1983Linear,Zhou2004,Roberts2008,1985Relation}. As the simplest and monotonic form of mapping, linear relationships offer high interpretability and intuitive understanding. This characteristic provides key advantages for sEMG-driven prosthetic hand control: it significantly simplifies the modelling between signals and outputs, reducing system complexity, and it enables users to more easily understand and operate the control logic, thereby enhancing intuitiveness and usability.
\begin{figure}[htbp]  
	\centering  
	\includegraphics[width=0.8\linewidth]{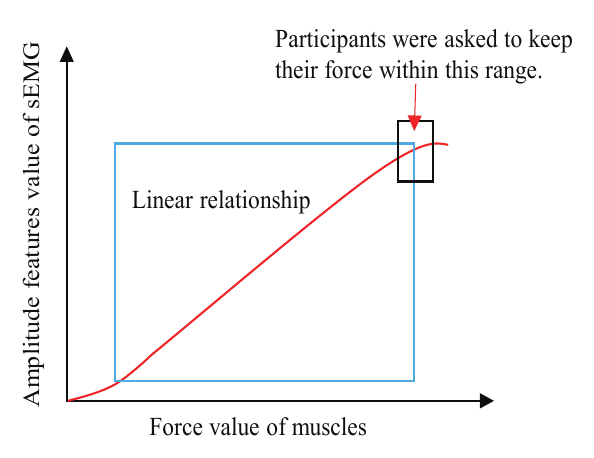} 
	\caption{Relationship between force values of muscles and amplitude features values of sEMG} 
	\label{fig:fig3} 
\end{figure} 

Mathematically, two points suffice to define a straight line. Inspired by this, we only record the sEMG corresponding to maximal finger extension and flexion, labelling the associated forces as +1 or -1, and fit the relationship between sEMG and muscle force using a near-linear model. This approach avoids the need for additional kinematic or kinetic data collection, while fully exploiting the inherent near-linear physiological relationship, allowing the mapping between sEMG and muscle force to be captured in a simple and rational manner. Consequently, the proposed method maintains prediction accuracy while achieving efficient and interpretable modelling, further enhancing the feasibility and clinical applicability of prosthetic hand control.

Specifically, we employed maximum voluntary contraction (MVC) to record the corresponding peak finger force. During isometric contraction, the forearm muscles responsible for MVC can generate maximal force while maintaining a constant muscle length, and the associated sEMG signals exhibit clear and recognisable features, allowing reliable recording and scaling ~\cite{1983Skeletal}. Although the relationship between muscle force and sEMG is not strictly linear under MVC, the recorded signals still serve as a robust approximation of force output at the upper bound of the near-linear range (see \autoref{fig:fig3}).

\begin{figure}[htpb] 
	\centering  
	\includegraphics[width=0.8\linewidth]{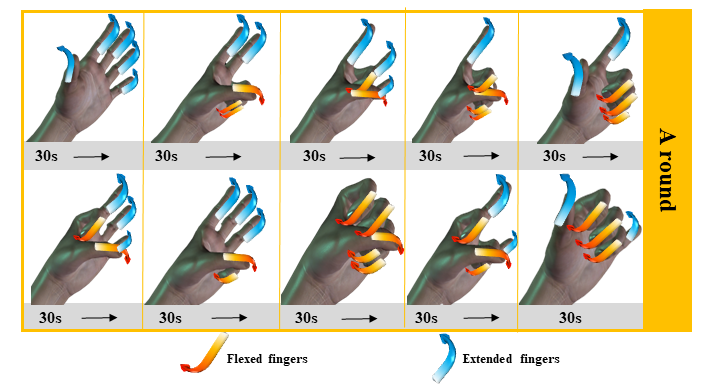} 
	\caption{List of 10 gestures in source gesture set and their force mode, participants hold each gesture for 30 seconds} 
	\label{fig:fig4}  
\end{figure} 

Accordingly, participants were instructed to perform maximum finger extension and flexion gestures to collect sEMG signals representing peak muscle strength. Based on this, we established two anchor points corresponding to the maximal extension and flexion forces during MVC, and assigned them labels of –1 and +1, respectively (see \autoref{fig:fig1}.A: Training), thereby defining the near-linear mapping between sEMG and muscle force.

\subsection{Proposed Method}

\begin{figure*}[htpb]  
	\centering
	\includegraphics[width=0.8\textwidth]{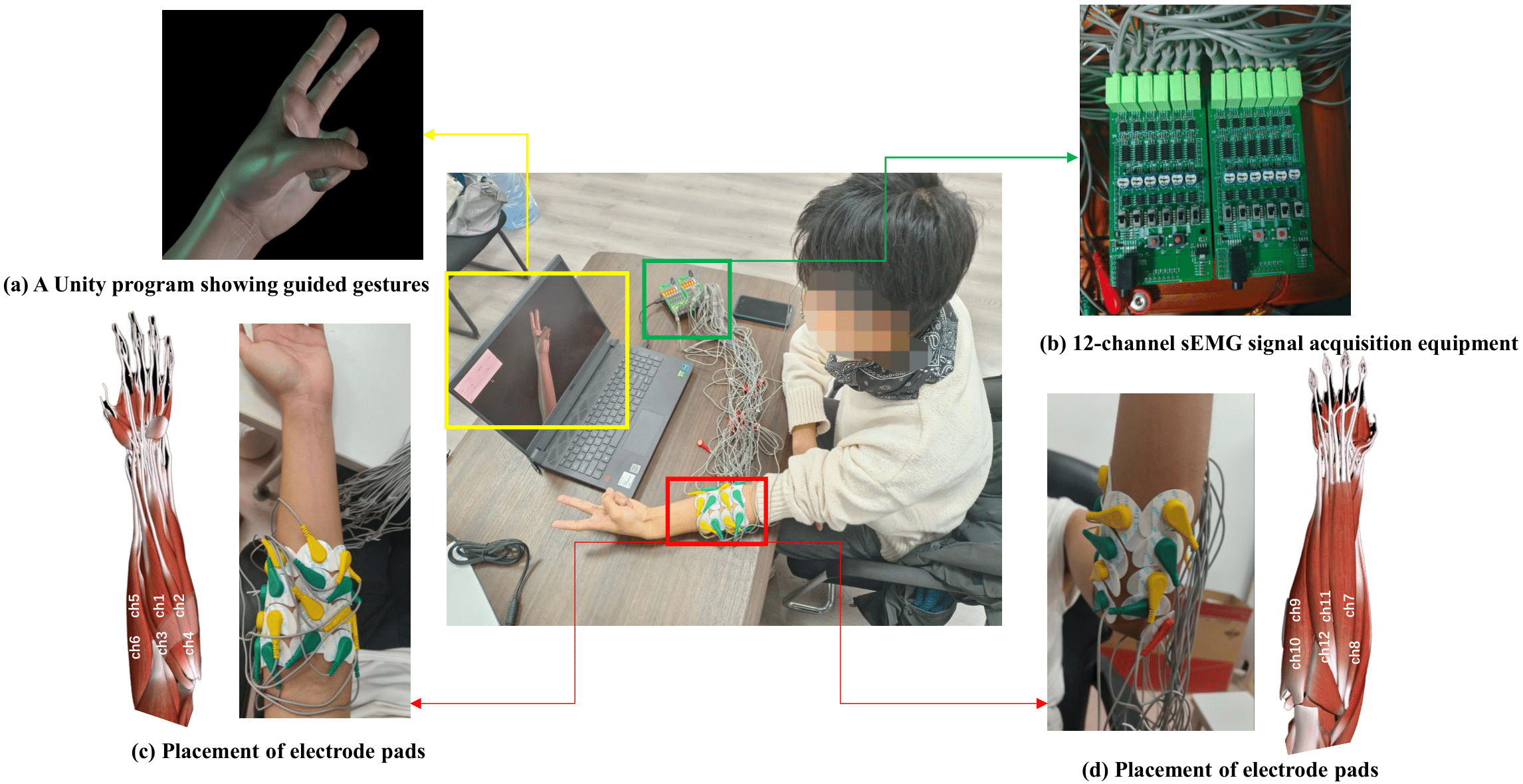}
	\caption{sEMG data were collected using a 12-channel sensor, covering as much of the forearm muscle groups as possible. The Unity 3D program was used to display guided gestures for participants in data collection}
	\label{fig:fig2}
\end{figure*}	

Our team previously proposed the Dendritic Neural Network (DD), which employs the Hadamard product to achieve nonlinear mapping while retaining linear computation, thereby providing strong representational capacity ~\cite{Luo2024,Liu2022a,Liu2022b,Liu2022,2020It}. However, as the network depth increases, this structure tends to introduce excessive nonlinearity, deviating from the near-linear characteristics required in this study. To address this issue, we introduce a residual mechanism into the DD framework and propose an improved two-layer ResDD model. The two-layer design aims to constrain the accumulation of nonlinear effects, thereby better aligning with the task of near-linear fitting.

The mathematical form is given by
\[
G = WX, \quad Y = G \circ X + G,
\]

where $X$ represents the input vector, $W$ represents the weight matrix, and $\circ$ represents the Hadamard product. The output can be decomposed as
\[
Y = G + \sigma(G, X), \quad \sigma(G, X) = (WX) \circ X.
\]

In this structure, the linear dominant term $G = WX$ preserves the primary linear relationship between input and output, while the nonlinear correction term $\sigma(G,X)$ compensates for details not captured by the linear mapping.
The residual mechanism provides two main benefits:

\begin{enumerate}
	\item \textbf{Preservation of linear dominance:} 
	When the nonlinear term $\sigma(G,X)$ remains small under parameter initialization or training constraints, the output approximates $Y \approx G$, thus maintaining overall linearity.
	
	\item \textbf{Progressive near-linear fitting:} 
	Since $\sigma(G,X) = (WX)\circ X$ directly depends on the weight matrix $W$, constraining the norm or gradient of $W$ explicitly controls the contribution of the nonlinear component, enabling gradual refinement from a linear backbone to nonlinear details.
\end{enumerate}

Overall, ResDD can be regarded as a composite mapping comprising a linear dominant term and a nonlinear correction term. Conceptually, this form is similar to a Taylor expansion: the first-order linear term serves as the backbone, while higher-order terms provide limited local correction. This structure preserves near-linear characteristics while enhancing flexibility and detail fitting.
\label{sec:guidelines}

\subsection{sEMG Data Acquisition and Feature Extraction}

\subsubsection{Data collection preparation}
In our study, twenty able-bodied participants (two females, eighteen males, aged 19.5 ± 1.05 years) provided informed consent to participate in the study protocol. All participants were right-handed and reported no history of neurological or muscular diseases. sEMG signals were recorded from the left forearm of each participant. The study protocol was approved by the ethics review board of Zhengzhou University (62303423) and adhered to the Declaration of Helsinki and relevant policies in China.

We designed ten gestures as source gesture sets as shown in \autoref{fig:fig4}. The standards are outlined in the supplementary document, under the section on gesture design standards.

\subsubsection{Data acquisition process}

The sEMG signals were acquired using a 12-channel surface electromyography system at a sampling rate of 1 kHz. The acquisition procedure is depicted in \autoref{fig:fig2}, and a comprehensive description is provided in the supplementary materials.

\begin{table}[htpb]  
	\centering  
	\small 
	\captionsetup{font=bf} 
	\caption{\footnotesize Selected Features with a Monotonic Relationship}\label{table1}  
	\begin{tabular}{@{}cclll@{}}  
		\bottomrule \hline \bottomrule 
		\begin{tabular}[c]{@{}c@{}}monotonic linear feature\end{tabular} & \multicolumn{4}{c}{Formula} \\ \midrule  
		\textbf{Root Mean Square (RMS)} & \multicolumn{4}{c}{\textbf{\small $\sqrt{\frac{1}{N} \sum_{i=1}^{N} x_i^2}$}} \\   
		\addlinespace[0.5ex] 
		\textbf{Mean Absolute Value (MAV)} & \multicolumn{4}{c}{\textbf{{\small $\frac{1}{N} \sum_{i=1}^{N} |x_i|$}}} \\    
		\addlinespace[0.8ex]  
		\textbf{Variance (VAR)} & \multicolumn{4}{c}{\textbf{{\small $\frac{1}{N-1} \sum_{i=1}^{N} (x_i - \bar{x})^2$ }}} \\  
		\addlinespace[0.8ex]  
		\textbf{Standard Deviation (SD)} & \multicolumn{4}{c}{\textbf{{\small $\sqrt{\frac{1}{N-1} \sum_{i=1}^{N} (x_i - \bar{x})^2}$}}} \\  
		\addlinespace[0.8ex]  
		\textbf{Integral (INT)} & \multicolumn{4}{c}{\textbf{{\small $\sum_{i=1}^{N} |x_i|$}}} \\  
		\addlinespace[0.5ex]  
		\textbf{Wavelength (WL)} & \multicolumn{4}{c}{\textbf{{\small $\sum_{i=1}^{N-1} |x_{i+1} - x_i|$}}} \\  
		\addlinespace[0.5ex]  
		\textbf{\begin{tabular}[c]{@{}c@{}} Difference Absolute Standard \\ Deviation Value (DASDV)\end{tabular}} & \multicolumn{4}{c}{\textbf{{\small $\sqrt{\frac{1}{N-1} \sum_{i=1}^{N-1} (x_{i+1} - x_i)^2}$}}} \\  
		\textbf{\begin{tabular}[c]{@{}c@{}}Difference Absolute Mean \\ Value  (DAMV)\end{tabular}} & \multicolumn{4}{c}{\textbf{{\small $\frac{1}{N-1} \sum_{i=1}^{N-1} |x_{i+1} - x_i|$ }}} \\  \bottomrule 
	\end{tabular}  
\end{table}

\subsubsection{Signal Preprocessing and Feature Extraction}

For each participant, the three sets of recorded signals were merged to construct a comprehensive dataset, which was subsequently pre-processed. The pre-processing~\cite{DeLuca2010,Vaziri} pipeline included DC offset removal~\cite{Merletti2020}, band-pass filtering within the 10–450 Hz range~\cite{DeLuca2010,2022A,2021Sheikhahmadi,8786584}, and notch filtering at 50 Hz~\cite{Piskorowski2013}. The signals were then full-wave rectified, followed by segmentation using a sliding window with a length of 200 ms and a step size of 50 ms(see \autoref{fig:fig5}), and a comprehensive description is provided in the supplementary materials.

To extract informative components from sEMG and suppress interference, feature extraction was applied to the filtered signals. Conventional features include time-, frequency-, and time–frequency-domain measures~\cite{Phinyomark2012,ZardoshtiKermani1995}. In this study, we focused on exploiting the monotonic near-linear relationship between sEMG and muscle force. Accordingly, eight amplitude-based time-domain features were extracted from each of the 12 channels: Root Mean Square (RMS), Mean Absolute Value (MAV), Variance (VAR), Standard Deviation (SD), Integral (INT), Wavelength (WL), Difference Absolute Standard Deviation Value (DASDV), and Difference Absolute Mean Value (DAMV). These features constitute the main representation of sEMG data in this paper, derivations of the feature matrices are provided in the supplementary materials.

As indicated in \autoref{table1},  we can infer that the values of these features have a monotonically relationship with the values of the original signal sequence. Thus, even after signal scaling, this monotonic relationship is retained in the feature values. Extracting features channel by channel also helps mitigate variability from electrode placement, yielding a robust dataset for muscle force analysis.

\section{Experimental setup}
\label{Experimental setup}
To systematically evaluate the effectiveness of the proposed ResDD model, three representative models were selected, covering a spectrum from strictly linear to highly nonlinear. The core formulas of these four models are shown in the \autoref{table2}. This design not only tests whether ResDD can improve performance while retaining linear dominance with controlled nonlinear correction, but also highlights its balance between interpretability and representational power.Details of the model configurations and experimental settings are provided in the supplementary document.

\begin{table}[htbp]  
	\centering  
	\small 
	\renewcommand{\arraystretch}{1}  
	\captionsetup{font=bf}  
	\caption{\footnotesize Core Formulas of Different Models} \label{table2}  
	\begin{tabular}{ccc}  
		\toprule
		Type & Model & Core Formula \\  
		\midrule  
		\multirow{2}{*}{Near-linear}  
		& ResDD & $G=WX,\; Y = G \circ X + G$ \\  
		\cmidrule(lr){2-3}  
		& LN    & $Y = WX$ \\  
		\midrule  
		\multirow{2}{*}{Nonlinear}  
		& MLP   & $Y = \text{ReLU}(WX+b)$ \\  
		\cmidrule(lr){2-3}  
		& CNN   & \shortstack[l]{$C = \text{ReLU}[\text{Conv}_{2D}(X)],\; P = \text{Pool}(C)$ \\ $Y = \text{ReLU}(WP+b)$} \\  
		\bottomrule  
	\end{tabular}  
\end{table}

\subsection{Verifying whether the near-linear model can learn sufficient information}
To verify the effectiveness of the near-linear model ResDD in learning sEMG features and decoding ability, we designed an offline experiment, using force direction classification as the benchmark for model learning capacity. In this task, output labels correspond to the magnitude and direction of finger force: positive for flexion, negative for extension, and zero as the threshold for direction classification, also used in calculating accuracy. Four models—ResDD, LN, MLP, and CNN—were trained for comparison. The area under the ROC curve (AUC) was used as the performance metric, with ROC curves drawn to visualise classification performance (see \autoref{fig:fig9}). By combining all participants’ outputs, five output sets (L1–L5) were generated for each model, and based on these, AUC values for different fingers were calculated (see \autoref{table 3}).

\section{Experimental Validation}
\label{Results}

\begin{figure*}[htpb]
	\centering
	\includegraphics[width=0.78\textwidth]{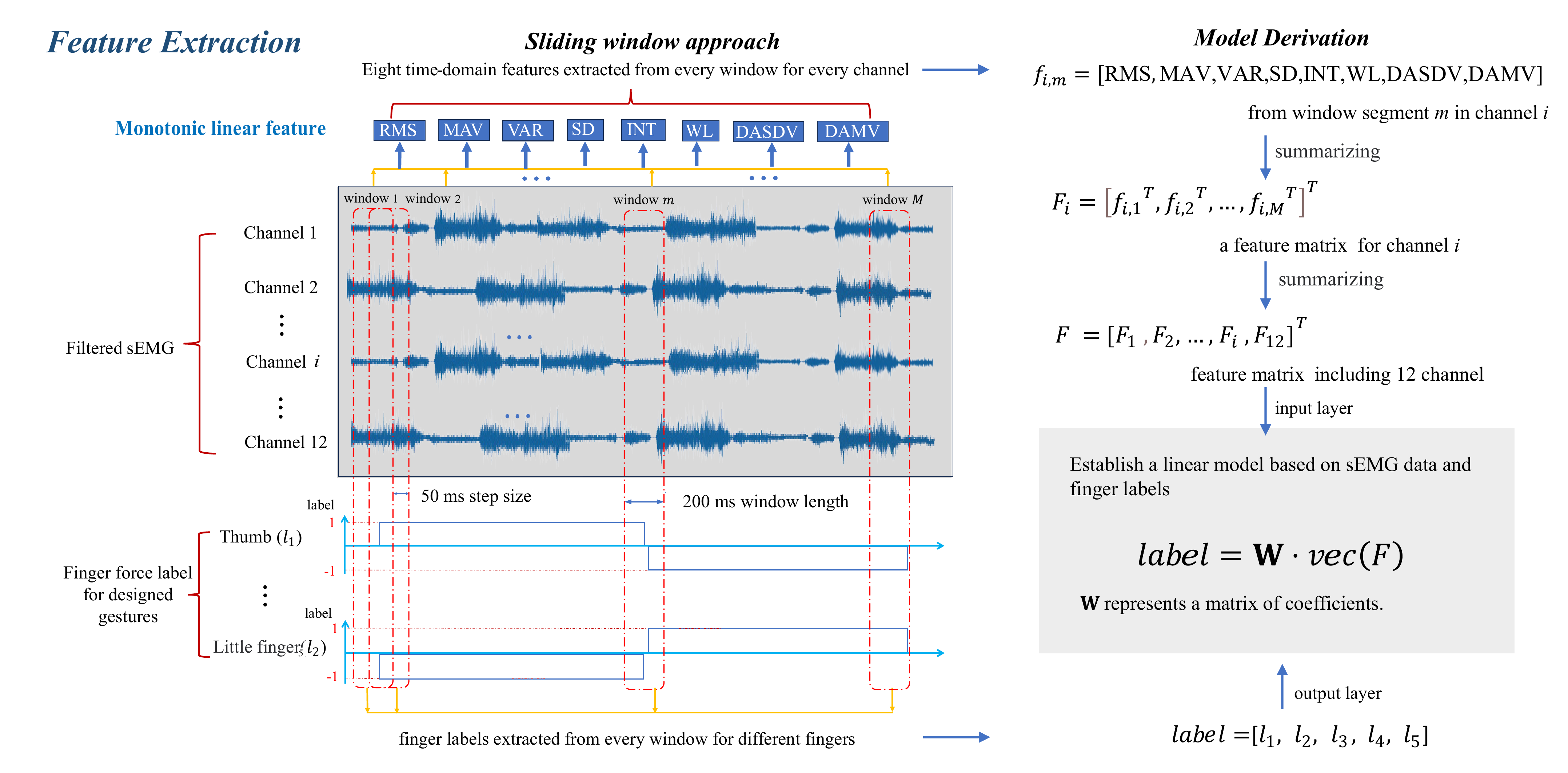}
	\caption{Feature Extraction. For filtered sEMG signals and finger force labels, a sliding window approach is used to extract eight time-domain features from each window across all channels. Model Derivation. A $12 \times M \times 8$ input matrix is constructed as input matrix, where 12 represents the channels, $ M $ represents the number of segments, and 8 monotoniclinear features extracted from each segment. The output matrix contains finger force labels for the five fingers. This setup is used to train models that predicts finger force labels based on the extracted input features, enabling the decoding of continuous force and velocity information}
	\label{fig:fig5}
\end{figure*}

\begin{figure}[htbp]
	\centering
	\includegraphics[width=0.8\linewidth]{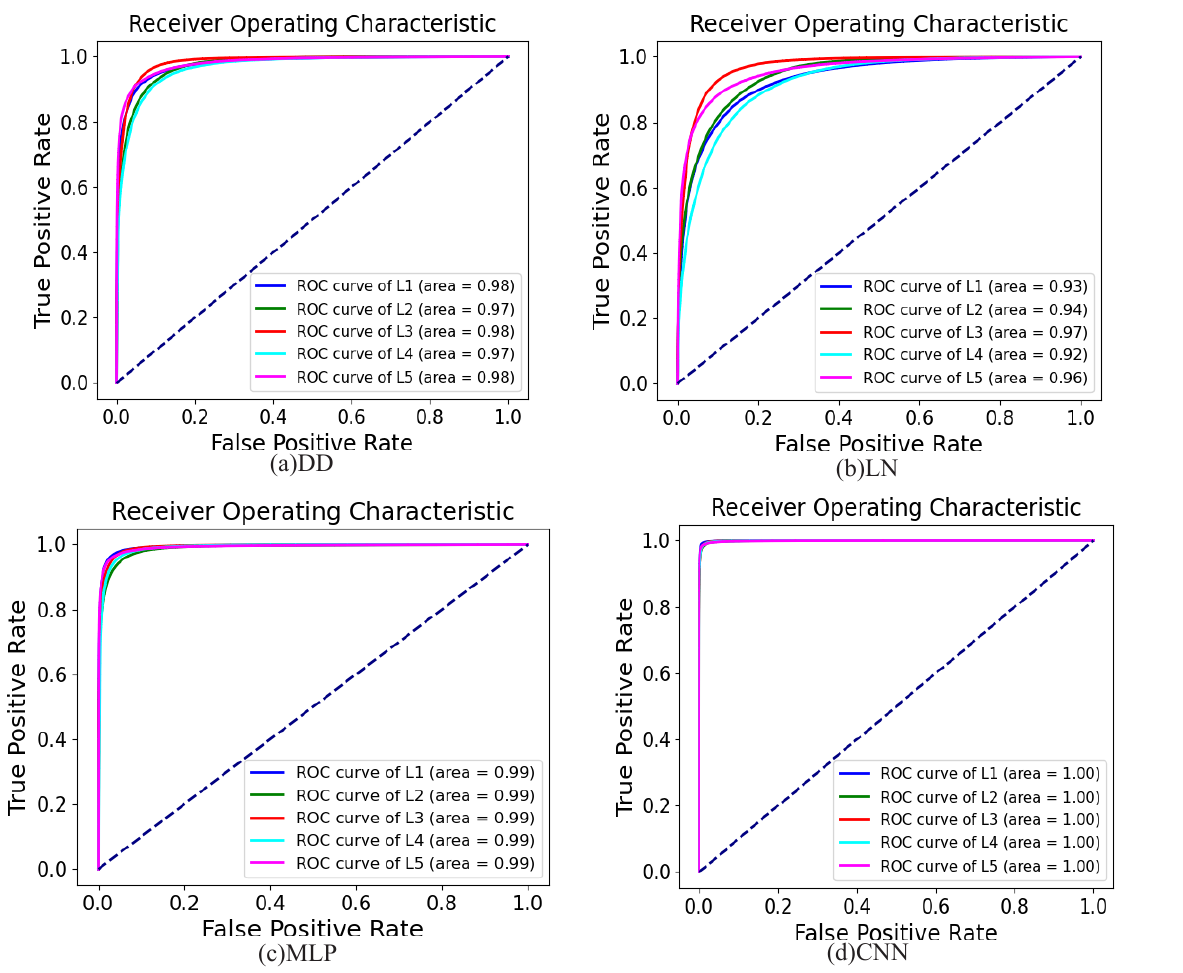}
	\caption{Visualize model performance for force direction classification}
	\label{fig:fig9}
\end{figure}

\begin{figure}[htbp]  
	\centering  
	\includegraphics[width=0.8\linewidth]{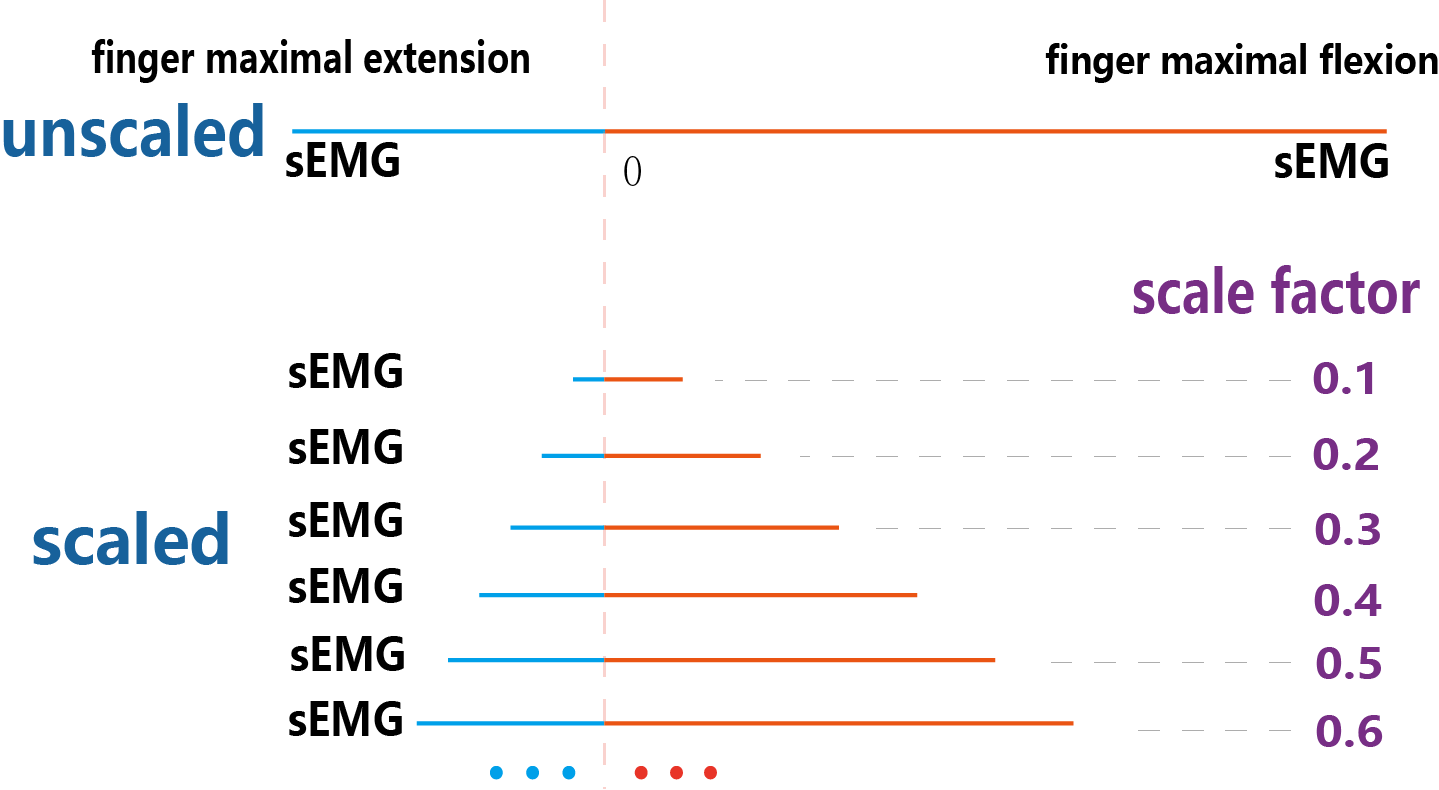}
	\caption{Scaling process for sEMG amplitudes in different directions (extension and flexion)} 
	\label{fig:fig7}
\end{figure}  	

\begin{table}[htbp] 
	\small 
	\centering 
	\captionsetup{font=bf} %
	\caption{\footnotesize Offline Analyses Results}\label{table 3}  
	\begin{tabular}{ccccc}  
		\bottomrule\hline\bottomrule
		Output & Method & \begin{tabular}[c]{@{}c@{}}Area Under the\\ Curve (AUC)\end{tabular} & \begin{tabular}[c]{@{}c@{}}Standard\\ Error (SE)\end{tabular} & Accuracy \\ \midrule  
		\multirow{4}{*}{L1} & ResDD & 0.977887 & 0.000449 & 92.22\% \\  
		& LN & 0.929772 & 0.000804 & 85.01\% \\  
		& MLP & 0.993835 & 0.000250 & 96.63\% \\  
		& CNN & 0.999411 & 0.000073 & 99.15\% \\ \midrule  
		\multirow{4}{*}{L2} & ResDD & 0.972789 & 0.000552 & 90.84\% \\  
		& LN & 0.942453 & 0.000798 & 86.07\% \\  
		& MLP & 0.988339 & 0.000382 & 94.50\% \\  
		& CNN & 0.998866 & 0.000113 & 98.66\% \\ \midrule  
		\multirow{4}{*}{L3} & ResDD & 0.982602 & 0.000398 & 93.79\% \\  
		& LN & 0.968013 & 0.000541 & 91.45\% \\  
		& MLP & 0.992689 & 0.000272 & 96.18\% \\  
		& CNN & 0.999116 & 0.000089 & 98.85\% \\ \midrule  
		\multirow{4}{*}{L4} & ResDD & 0.967460 & 0.000506 & 90.94\% \\  
		& LN & 0.919576 & 0.000812 & 84.78\% \\  
		& MLP & 0.989969 & 0.000292 & 95.55\% \\  
		& CNN & 0.999032 & 0.000086 & 98.84\% \\ \midrule  
		\multirow{4}{*}{L5} & ResDD & 0.980862 & 0.000529 & 93.94\% \\  
		& LN & 0.955773 & 0.000797 & 90.28\% \\  
		& MLP & 0.992517 & 0.000351 & 96.65\% \\  
		& CNN & 0.998840 & 0.000131 & 98.67\% \\ \bottomrule  
	\end{tabular}  
	\caption*{*This experiment focuses on the model's ability to learn useful representations, rather than on mere accuracy metrics.}
\end{table} 

The results show that ResDD can extract richer feature information and has stronger decoding ability than the linear model LN, while its classification performance is close to that of the nonlinear models (MLP and CNN). This confirms the effectiveness and advantage of ResDD in sEMG signal decoding.

\subsection{Verifying the model’s ability to fit the near-linear relationship between EMG  and force}
\begin{figure*}[htbp]
	\centering
	\includegraphics[width=0.8\textwidth]{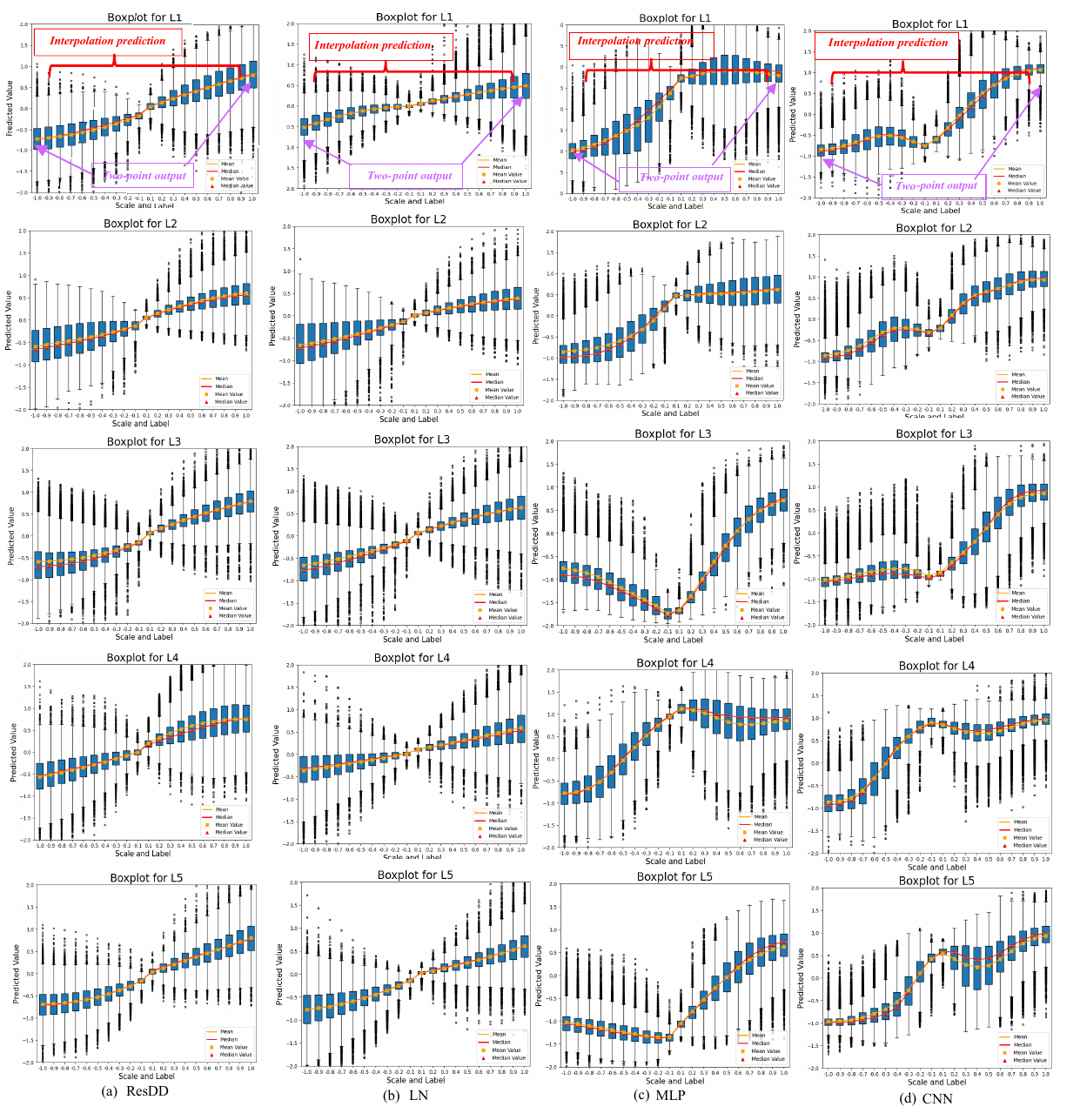}
	\caption{Visualization of Interpolation Results. The figure displays representative interpolation results from one subject's result. The abscissa represents the scale factor applied to the sEMG data, where positive values indicate finger flexion, and negative values indicate finger extension. The two ends of each image represent the outputs at the two extreme points (-1 and 1), while the middle section shows the interpolated outputs from the models. From L1 to L5, the labels correspond to the pinky to the thumb. Results Analysis. Panels (c) and (d) highlight errors where monotonicnear-linear fitting could not be achieved. The failure of these models for monotonic near-linear interpolation shows the limitations of complex models in maintaining linearity and monotonicity relation between predicted force and sEMG}
	\label{fig:fig10}
\end{figure*}

To further examine whether the model can properly fit the near-linear relationship between sEMG and force, we carried out an interpolation analysis to explore the internal properties of the trained models. Specifically, we extracted scaled surface EMG features from the test set (scaling process shown in \autoref{fig:fig7}) and fed them into the four trained models to simulate real application scenarios. If the fitted curve showed non-monotonic regions, the model was deemed to have failed in fitting, indicating a lack of intuitive control. \autoref{fig:fig10} presents interpolation results from one participant, while results for the remaining 19 participants are provided in the supplementary material. A summary of failure counts for the four models is shown in \autoref{table 4}.

The results indicate that both ResDD and the linear model LN can effectively fit the approximate linear relationship between EMG and force. In contrast, MLP and CNN showed more instances of fitting failure. These failures could result in less intuitive user experiences in practical control, requiring users to invest more time in learning and adaptation. This demonstrates that ResDD is better suited for the control framework proposed in this study. Furthermore, the results validate that the ResDD model can accurately capture the near-linear relationship between EMG and force.

Overall, ResDD outperforms the linear model LN in terms of decoding accuracy, and overcomes the problem of lack of intuitive control caused by excessive nonlinearity in MLP and CNN. While maintaining an approximate linear relationship, it can achieve higher accuracy and thus has higher practical value. Therefore, ResDD achieves the best balance among accuracy, interpretability, and system applicability, and becomes the most suitable choice in the proposed continuous prosthetic hand control framework in this paper.

\begin{table}[htbp]
	\small 
	\centering 
	\captionsetup{font=bf} %
	\caption{\footnotesize Statistics Analysis in Fitting Result}\label{table 4} %
	\begin{tabular}{ccccc}
		\bottomrule  \hline  \bottomrule
		Network             & ResDD          & LN          & MLP        & CNN        \\ \hline
		Error Times         & 7          & 9           & 69         & 58         \\ \hline
		Correct rate        & 93\%        & 91\%        & 31\%       & 42\%       \\          \bottomrule 
	\end{tabular}
\end{table}

\subsection{Verifying the feasibility of the proposed method in practical applications}
\subsubsection{sine wave tracking experiment}

To evaluate the real-time control of finger force magnitude and direction in sEMG-driven prosthetic hands, we designed a sine wave tracking experiment. Four subjects wore sEMG acquisition devices, observed a static sine wave with amplitude 1 on a screen, and attempted to track the target waveform by adjusting the direction and magnitude of their finger force. The collected sEMG  were processed and features extracted in real time, then input into the trained models. Model outputs were constrained to the range [-1, 1] and plotted in real time as control curves. A well-performing model would generate a control curve closely matching the target sine wave. 

\begin{figure}[htbp]
	\centering
	\includegraphics[width=0.8\linewidth]{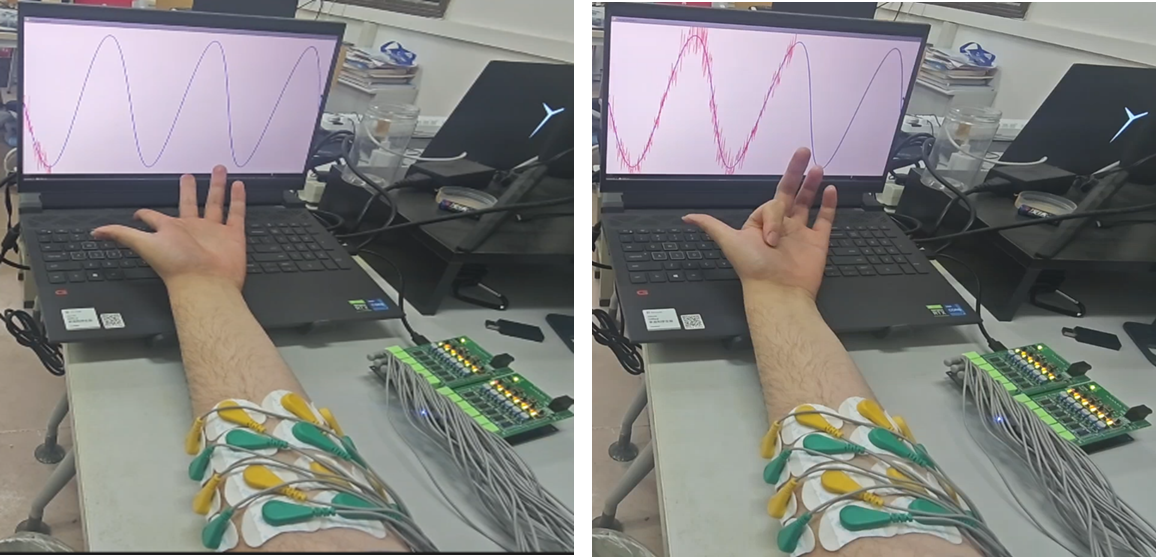}
	\caption{Sine-wave tracking experiment}
	\label{fig:fig11}
\end{figure}

\begin{figure}[htbp]
	\centering
	\includegraphics[width=0.82\linewidth]{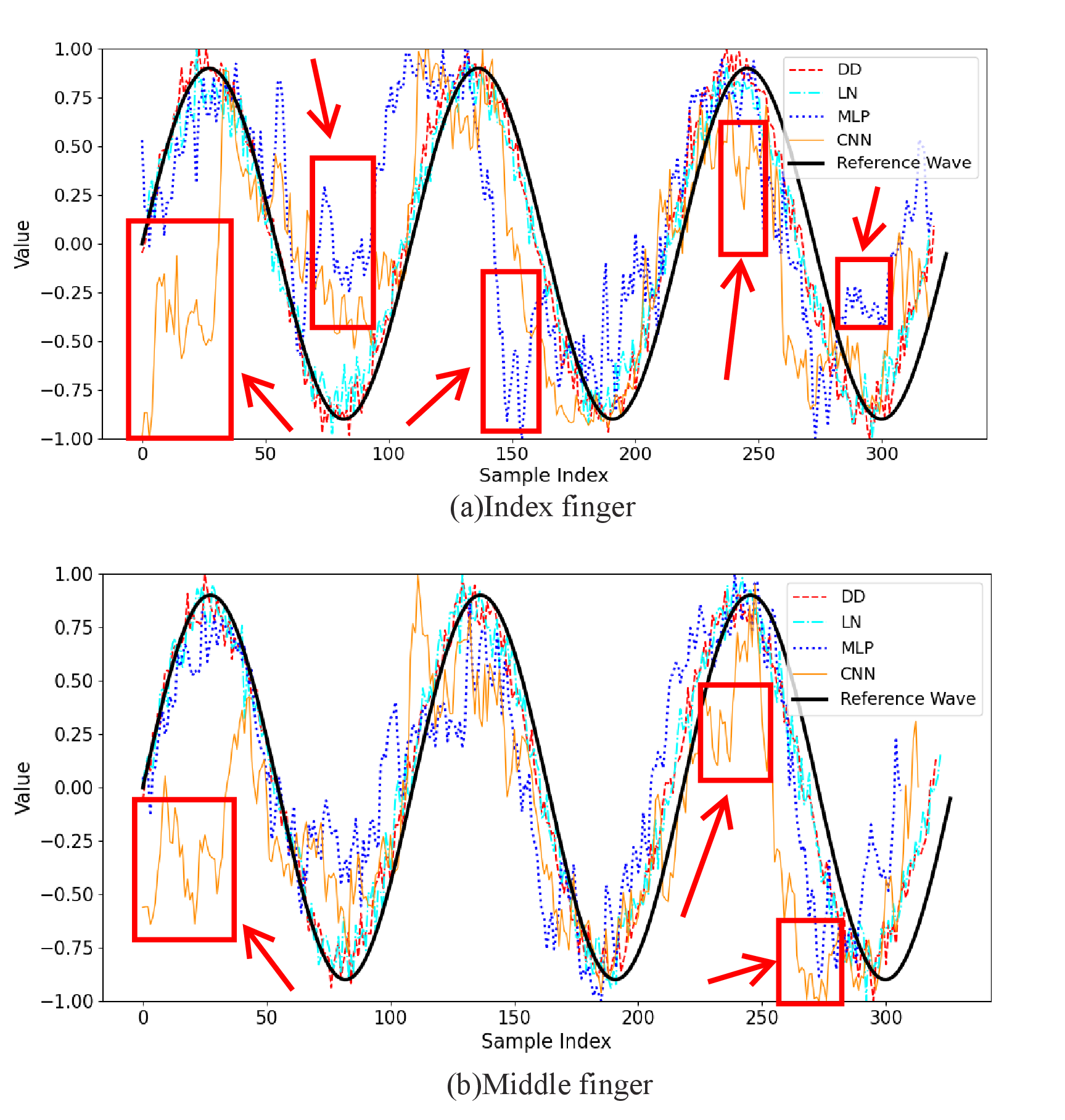}
	\caption{Waveform diagram of the index and middle finger sine-wave tracking experiment, different color curves correspond different models}
	\label{fig:fig12}
\end{figure}

\begin{figure}[htbp]
	\centering
	\includegraphics[width=0.9\linewidth]{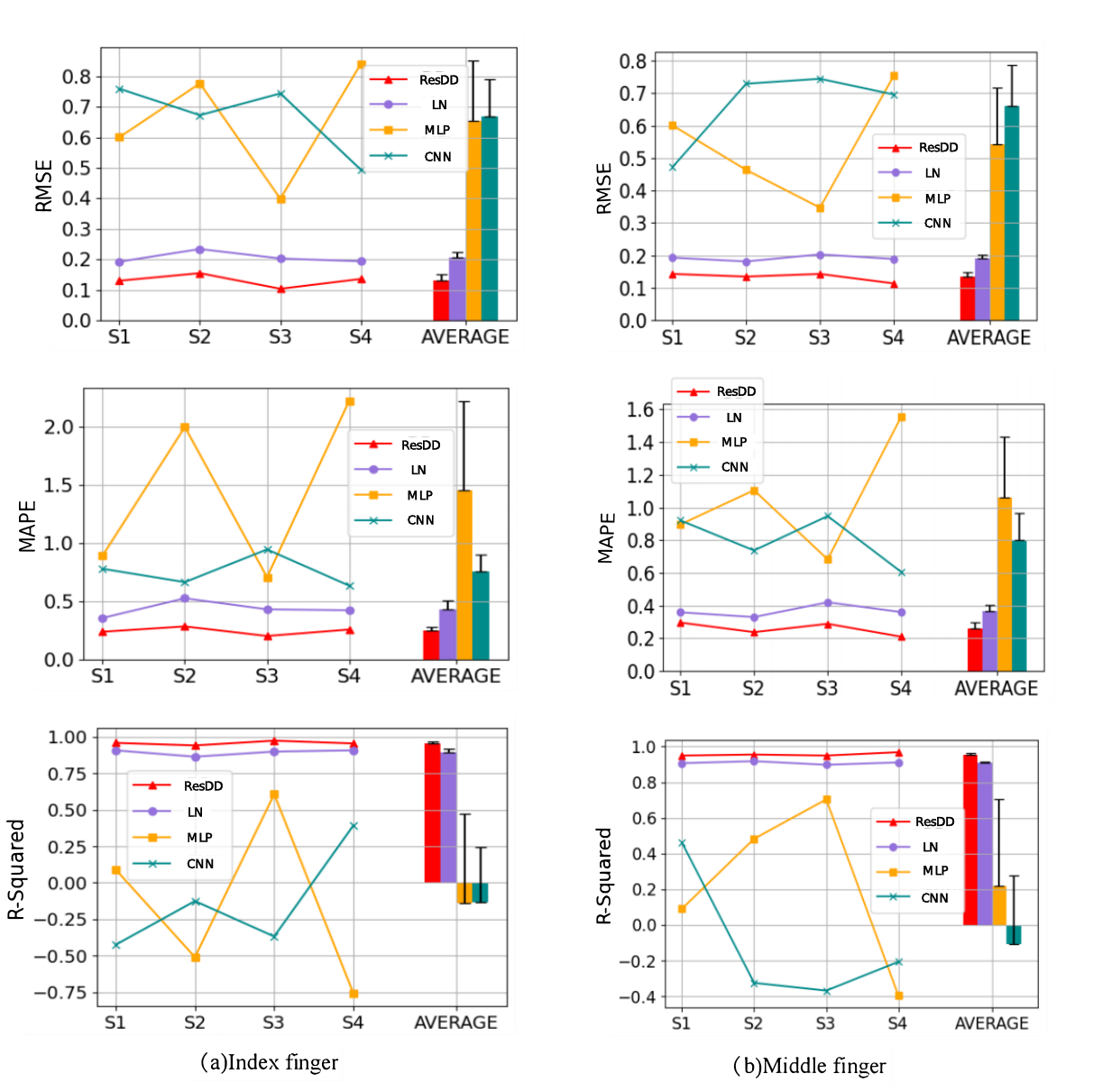}
	\caption{Performance of RMSE, MAPE, and R-squared metrics in an online sine-wave tracking experiment conducted with four subjects. The error bars represent the standard deviation (SD) across subjects for each metric}
	\label{fig:fig13}
\end{figure}

The experiment adopted a single-finger control design, meaning that only the output signal of the corresponding finger was used to generate the curve in each trial. This choice reduced the difficulty of controlling both force direction and magnitude, allowing the decoding and control performance of the models to be assessed more directly. The index and middle fingers—commonly used in daily activities—were selected as the focus. To ensure fairness, subjects were not informed of which model type they were using during the experiment. The experimental setup is illustrated in \autoref{fig:fig11}, and tracking results for the index and middle fingers are shown in \autoref{fig:fig12}. By comparing the real-time output curves with the target sine wave, the models' realtime force control ability can be visually assessed. In addition, we calculated the mean root mean square error (RMSE), mean absolute percentage error (MAPE), and correlation coefficient (R²) to provide more precise performance evaluation. The statistical results for these metrics are presented in \autoref{fig:fig13}. 

Together, Figures~\ref{fig:fig12} and~\ref{fig:fig13} show that traditional non-linear models (MLP and CNN) performed poorly in controlling both force magnitude and direction, whereas the near-linear models (ResDD and LN) achieved superior stability and accuracy.

\subsubsection{Real-Time Control Demostration}
\begin{figure}[htbp] 
	\centering  
	\includegraphics[width=0.8\linewidth]{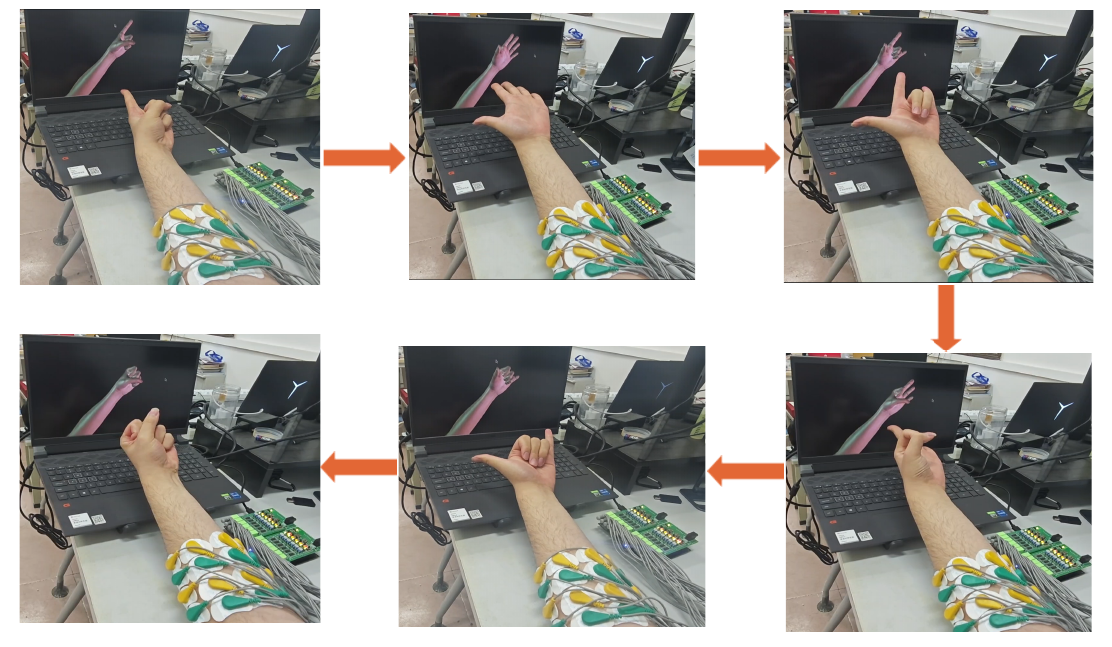}
	\caption{Real-time control of virtual hands}  
	\label{fig:fig14}  
\end{figure}

To validate the effectiveness of the proposed method in practical applications, we designed an online real-time control experiment, using the ResDD model to control all five fingers of virtual hands (see Fig.~\ref{fig:fig14}). To preserve the monotonic near-linear sEMG–force relationship defined in \autoref{eq2}, the output labels were used to proportionally control finger force and joint angular acceleration, with positive and negative labels distinguishing flexion and extension:

\begin{equation}\label{eq4}
	F = k_F \cdot label
\end{equation}

where $F$ denotes the actual output force, \textit{label} is the model’s force output, and $k_F$ is a userspecific scaling parameter that adjusts the actual output according to individual muscle strength and usage requirements. For example, in rehabilitation training, $k_F$ can be set to a smaller value to enable slow and controlled movements, whereas in daily prosthetic use, $k_F$ may be increased to ensure sufficient force output.

Similarly, the angular acceleration of the fingers is defined as follows:

\begin{equation}\label{eq5}
	\alpha = k_\alpha \cdot \text{label}
\end{equation}
where $k_\alpha$ is a scaling factor used to convert the force label into the actual angular acceleration. Combined with the physical relationship between angular velocity and acceleration:

\begin{equation}\label{eq6}
	\omega = \omega_0 + \alpha \cdot t
\end{equation}

real-time angular velocity and continuous gestures can be decoded. By adjusting $k_F$ and $k_\alpha$, the system can flexibly adapt to different user requirements, enabling personalized regulation between joint angular acceleration and output force.

In addition, two supplementary videos demonstrate real-time virtual five-finger control based on the proposed method. The results show that the system can achieve natural, intuitive, and flexible gesture estimation without relying on kinematic or force sensors, confirming that prosthetic hand control based on predefined labels and monotonic near-linear interpolation can accurately capture and translate user intent.

\section{Discussion}
\label{Discussion}
This paper proposes a continuous prosthetic hand control method based on sEMG, which does not require complex motion or force sensor calibration. By exploiting the near-linear relationship between electromyography and force under specific conditions, the ResDD model is employed for fitting, thereby enabling the decoding of electromyographic signals for control. This approach simplifies system structure, maintains monotonicity, and provides interpretability and intuitiveness, enabling users to receive feedback and make real-time adjustments more easily. It reduces modeling complexity, allows users to grasp the control logic more readily, and significantly enhances system practicality and usability. Specifically, the method adopts the principle of “two points determine a straight line”: sEMG signals from the hand’s maximum flexion and extension states are labeled as 1 and -1, respectively, serving as the basis for system calibration. The proposed ResDD model then effectively fits the near-linear sEMG–force relationship, eliminating the cumbersome calibration steps required by traditional methods and improving deployment efficiency while maintaining functional integrity.

To systematically evaluate ResDD, we compared it with linear (LN) and nonlinear models (MLP and CNN). As representatives of fully connected and convolutional structures, MLP and CNN have strong nonlinear expression capabilities and, according to the universal approximation theorem, can theoretically approximate any nonlinear function. However, if these models fail to capture the approximate linear mapping in this task, more complex models (e.g., RNN or Transformer) may introduce stronger nonlinearities, increasing control complexity and reducing user convenience. Therefore, using MLP and CNN as nonlinear benchmarks suffices to highlight the limitations of nonlinear methods in this context.

An offline classification experiment was designed to verify whether ResDD can effectively learn sEMG characteristics and achieve accurate decoding. Evaluating force-direction classification performance allowed us to determine whether the model captured sufficient key information. Results show that ResDD outperforms the linear model in feature extraction and decoding, and its performance is comparable to that of nonlinear models.

To assess whether the model accurately captures the near-linear sEMG–force relationship, we conducted interpolation analyses on the trained models. Visualizing the fitted curves revealed that both ResDD and LN maintained monotonic and approximately linear responses, enabling users to adjust actions in real time during closed-loop feedback and providing an intuitive, controllable interactive experience. In contrast, MLP and CNN produced curves with non-monotonic and nonlinear intervals, potentially causing discrepancies between control output and user expectation, thereby reducing intuitiveness and usability.

To evaluate practical feasibility, online real-time control and sine-wave tracking experiments were conducted. Sine-wave tracking demonstrated that ResDD achieved more accurate control responses, while online experiments confirmed that users could complete effective real-time interaction tasks. These results demonstrate the method’s practicality and superiority in real-world applications.

In conclusion, we propose a continuous prosthetic hand control method based on the approximate linear mapping between muscle contraction force and sEMG amplitude, significantly simplifying control. The method enables intuitive, continuous control while ensuring accuracy and stability, reducing the development and usage barriers of traditional sensor-based solutions. This work introduces a new design concept for electromyography-controlled prosthetics, offering strong practicality and scalability, and laying the foundation for future development of more natural and interpretable multi-joint control.

\begin{figure}[htbp] 
	\centering  
	\includegraphics[width=0.8\linewidth]{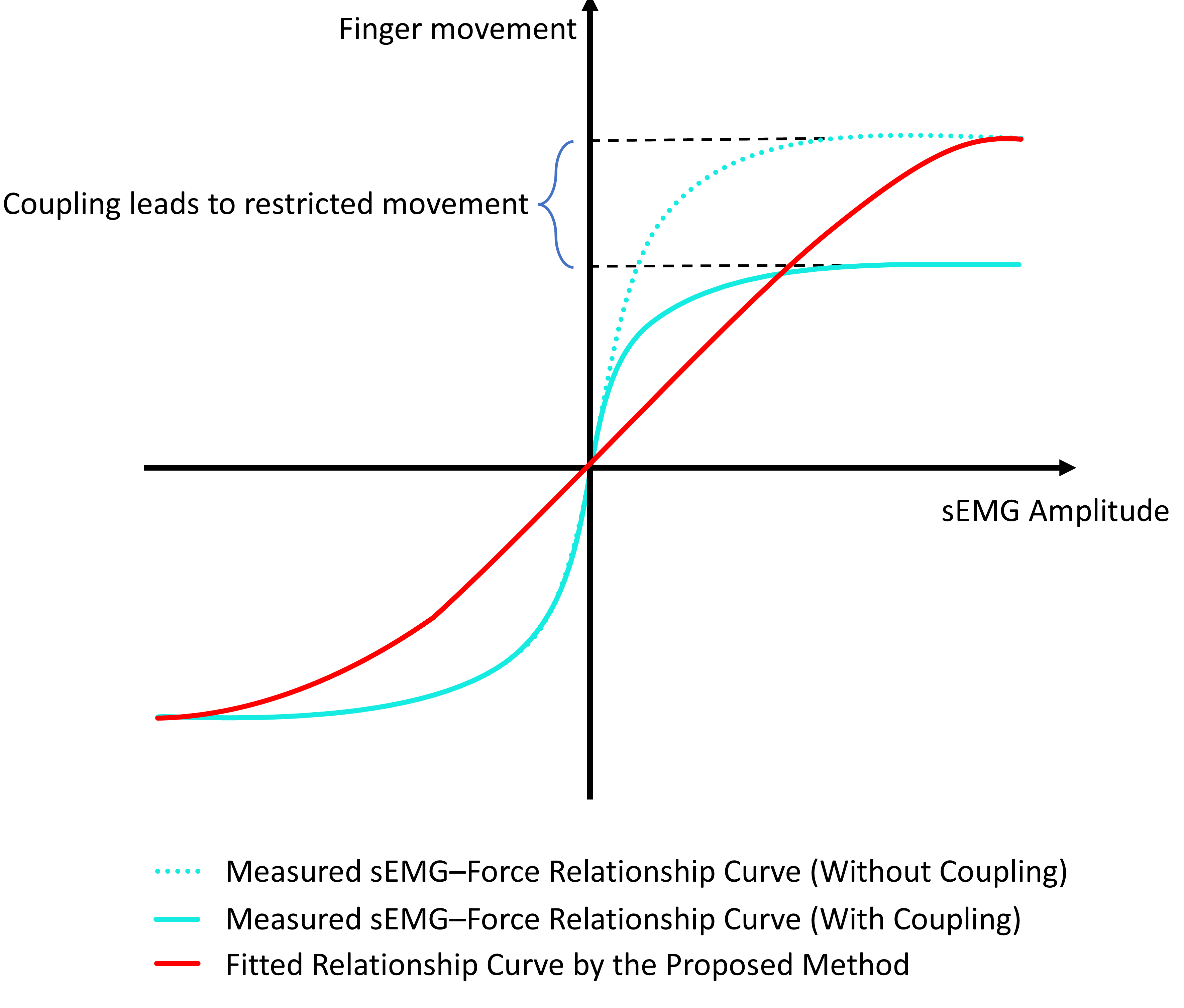}
	\caption{Schematic of the Near-Linear Fitted Curve with Finger Coupling}  
	\label{fig:fig15}  
\end{figure}

A limitation of this study is that all participants were non-disabled. Although previous studies indicate some similarity in muscle activation patterns between amputees and non-disabled individuals~\cite{2014Intuitive}, further verification with amputee participants is necessary. Notwithstanding this, the findings hold significant promise for clinical translation. The proposed method addresses a fundamental barrier in current regressively-based sEMG prosthetics: the dependency on external kinematic or dynamic sensors for calibration. By establishing a near-linear relationship between sEMG and force that can be defined simply by two extreme points (maximum flexion and extension), our ResDD model eliminates the need for complex sensor calibration routines. This core innovation holds direct and significant clinical promise. For amputee users, it could translate into a more practical and user-centric prosthetic system, characterized by a substantially simplified donning-and-go process, reduced hardware complexity and cost, and ultimately, a lower barrier to adoption. For amputee users, this could translate to a reduced cognitive burden and more reliable control in daily activities. Additionally, while finger coupling (activation of one finger triggering sEMG in others~\cite{2004Human,2000Quantifying}) was not fully addressed, the primary aim was to propose an intuitive, user-friendly control method (see \autoref{fig:fig15}). The approximate linear mapping established here allows users to perceive operational effects clearly during closed-loop feedback and make timely adjustments, achieving effective real-time control even in the presence of coupling, further enhancing intuitiveness and usability.

\section{Conclusion}
\label{Conclusion}

This paper proposes a continuous prosthetic hand control system based on sEMG , which eliminates the need for complex kinematic or force sensor calibration. Specifically, a simplified near-linear relationship between sEMG and force is established by recording the sEMG under two maximum voluntary contraction (MVC) states and determining a straight line between two points. This relationship is then fitted using the near-linear model, ResDD. The model’s classification performance is evaluated through offline experiments, demonstrating that ResDD can effectively learn sEMG features and decode them accurately. Further interpolation analysis visualizes the model’s internal mapping, confirming that ResDD maintains a near-linear relationship. Compared to nonlinear models like MLP and CNN, ResDD offers a more intuitive control mechanism, facilitating easier user interaction. Finally, online sine wave tracking and real-time control experiments validate the superiority and feasibility of the proposed method in practical applications, showing that fitting a simple near-linear relationship between sEMG and force is sufficient to achieve effective control.

\section*{Acknowledgment}

The authors are profoundly indebted to all the patients and volunteers who participated in this trial. Without their generosity, this work would not have been possible. Finally, we acknowledge the anonymous reviewers for their constructive comments.

The authors declare no competing interests.

\bibliographystyle{IEEEtran} 
\bibliography{sss1}

\end{document}